\documentclass[lettersize,journal]{IEEEtran}
\usepackage{amsmath,amsfonts}
\usepackage{algorithmic}
\usepackage{algorithm}
\usepackage{array}
\usepackage[caption=false,font=normalsize,labelfont=sf,textfont=sf]{subfig}
\usepackage{textcomp}
\usepackage{stfloats}
\usepackage{url}
\usepackage{verbatim}
\usepackage{graphicx}
\usepackage{hyperref}
\usepackage[autostyle]{csquotes}

\usepackage[inline]{enumitem}
\usepackage{graphicx}

\usepackage{booktabs}
\usepackage[table,xcdraw]{xcolor}
\usepackage{pifont}
\usepackage{amsmath}
\usepackage{multirow,makecell}
\usepackage[figuresright]{rotating}
\usepackage{longtable}
\usepackage{float}
\usepackage{caption, tabularx} 

\newcommand{\ProposedFrameworkMERRY}{\textit{MERRY}} 
\newcommand{\ProposedDatasetMERRY}{\textit{MERRY-Data}}
\newcommand{\Findings}{
(1) Training on synthetic datasets tends to reduce emotional consistency, whereas training on real-world datasets improves it;
(2) Existing models suffer from emotional templatization and simplification, exhibiting positive-bias and performance bottleneck in fine-grained negative emotions.
(3) Simple prompting method strengthens the weak models but constrains the strong ones, while simple fine-tuning method suffers from poor role generalization. 
} 

\begin{document}

\title{\ProposedFrameworkMERRY: Semantically Decoupled Evaluation of \underline{M}ultimodal \underline{E}motional and \underline{R}ole Consistencies of \underline{R}ole-Pla\underline{y}ing Agents}

\author{Zhenyu Wang, ~\IEEEmembership{Student Member,~IEEE,} Xiaofen Xing, ~\IEEEmembership{Member,~IEEE,} Yirong Chen, ~\IEEEmembership{Student Member,~IEEE,} Xiangmin Xu, ~\IEEEmembership{Senior Member,~IEEE}
\thanks{Manuscript updated January 31, 2026. (Corresponding authors: Xiaofen Xing, Xiangmin Xu.)}
\thanks{Z. Wang, X. Xing and Y. Chen are with the South China University of Technology, Guangzhou 510641, China (e-mail: ftzywang@mail.scut.edu.cn; xfxing@scut.edu.cn; eeyirongchen@mail.scut.edu.cn).}
\thanks{X. Xu is with the Foshan University, Foshan 528225, China, and also with the South China University of Technology, Guangzhou 510641, China (e-mail: xmxu@scut.edu.cn).}
}

\markboth{Journal of \LaTeX\ Class Files,~Vol.~14, No.~8, August~2021}%
{Shell \MakeLowercase{\textit{et al.}}: A Sample Article Using IEEEtran.cls for IEEE Journals}

\IEEEpubid{0000--0000/00\$00.00~\copyright~2021 IEEE}

\maketitle

\begin{abstract}
Multimodal Role-Playing Agents (MRPAs) are attracting increasing attention due to their ability to deliver more immersive multimodal emotional interactions. However, existing studies still rely on pure textual benchmarks to evaluate the text responses of MRPAs, while delegating the assessment of their multimodal expressions solely to modality-synthesis metrics.
This evaluation paradigm, on the one hand, entangles semantic assessment with modality generation, leading to ambiguous error attribution, and on the other hand remains constrained by the heavy reliance on human judgment.
To this end, we propose {\ProposedFrameworkMERRY}, a semantically decoupled evaluation framework for assessing \textbf{\underline{M}}ultimodal \textbf{\underline{E}}motional and \textbf{\underline{R}}ole consistencies of \textbf{\underline{R}}ole-pla\textbf{\underline{y}}ing agents.
This framework introduce five refined metrics for EC and three for RC. Notably, we transform the traditional subjective scoring approach into a novel bidirectional-evidence-finding task, significantly improving the human agreement of LLM-as-Judge evaluations.
Based on {\ProposedFrameworkMERRY}, we conduct extensive evaluations. 
Our empirical results primarily reveal that:  
{\Findings}
Codes and dataset are available\footnote{\url{https://github.com/Luffy966/MERRY}}.

\end{abstract}


\begin{figure*}[htbp]
	\centering
	\includegraphics[width=\textwidth]{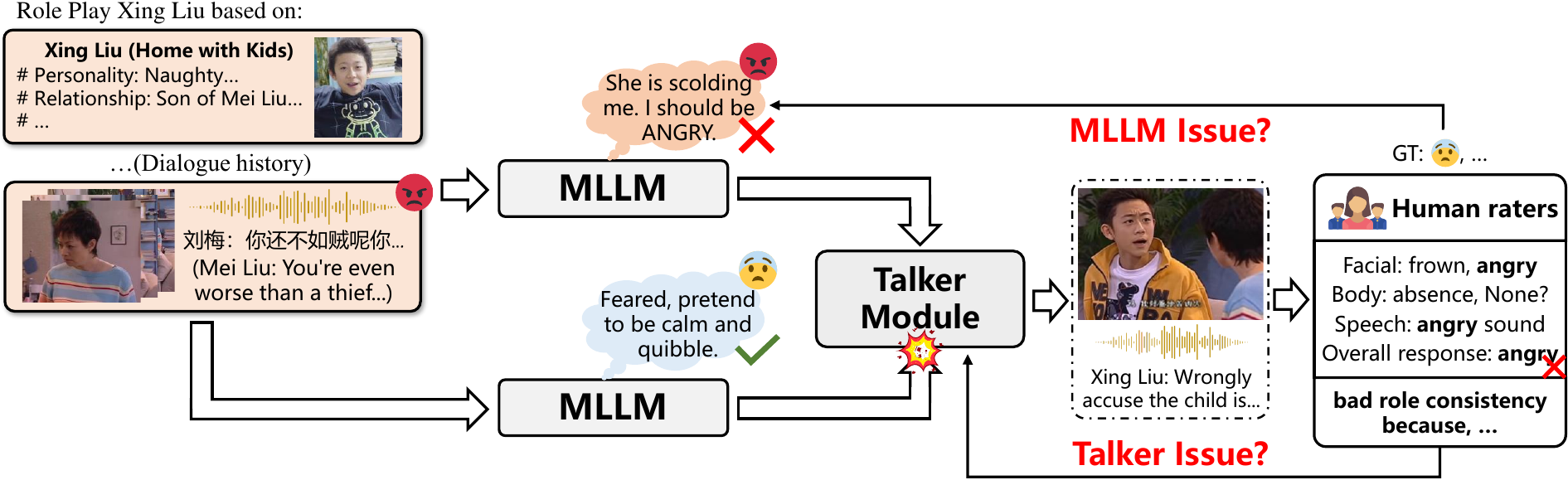} 
	\caption{Current end-to-end evaluations lead to ambiguous error attribution and strong dependence on human assessment. For instance, given a same video output and end-to-end human evaluation result, the issue can be ambiguously attributed to either MLLM (top) or Talker (bottom).}
	\label{fig:Illustration_of_question}
\end{figure*}

\section{Introduction}
\IEEEPARstart{M}{ultimodal} Role-Playing Agents (MRPAs) \cite{dai2025mmrolecomprehensiveframeworkdeveloping} have attracted growing attention for their ability to deliver more immersive interactions through multimodal emotional expressions, including facial expressions, body movements, and speech intonations, instead of pure textual contents.
However, existing studies \cite{zhang2025omnicharacterimmersiveroleplayingagents,chen2025emoavatar} still rely on textual role-playing benchmarks \cite{chen-etal-2024-socialbench, tu-etal-2024-charactereval} to evaluate the text content parts from responses of MRPAs, while leaving the assessment of their multimodal expressions to modality-synthesis metrics, typically via automatic and human evaluations.
As depicted in Figure~\ref{fig:Illustration_of_question}, this coupled evaluation paradigm, on the one hand, entangles semantic assessment with modality generation, thereby leading to ambiguity in issue attribution.
On the other hand, current modality-synthesis automatic metrics often fail to evaluate the underlying semantic-level dimensions, such as cross-modal consistency, emotional expressiveness and role-style consistency \cite{chen2025emoavatar,zhang2025multimodalempatheticresponsegeneration}.
Therefore, it exhibits a strong dependence on human judgment.

\IEEEpubidadjcol

To bridge this gap, we propose {\ProposedFrameworkMERRY}, which decouples the evaluation process into semantic and modality, and prioritizes the under-explored semantically decoupled evaluation on the following two dimensions.
\textbf{Emotional Consistency (EC)} typically refers to whether Role-Playing Agents (RPAs) can simulate the real and dynamic emotions of characters during dialogues \cite{feng-etal-2025-emocharacter}.
When there are multimodal emotional expressions, EC should further measures the degree of emotional consistency across different modalities. 
Psychological researches have found that human emotional expression is often presented through the collaboration of multiple modalities, and emotional consistency is maintained across different modalities \cite{Van_den_Stock2007-nb, Hollenstein-2014-emotionalconcordance}, which makes the aforementioned requirement for cross-modal emotional consistency have the rationality of human-likeness. 
Otherwise, inconsistent emotions across modalities will cause a severe sense of dissonance, which in turn significantly undermines the immersive experience.
Additionally, a response often comprises multiple utterances separated by punctuation, which may convey same or different emotions.
This suggests that emotional transitions consist of not only inter-turn, but also intra-turn, which is overlooked in previous research.
\textbf{Role Consistency (RC)} typically refers to how much RPAs can maintain the knowledge \cite{tu-etal-2024-charactereval} and personality \cite{wang-etal-2024-incharacter} of designated characters.
For multimodal dialogues, integrating multimodal expressions during the evaluation process is crucial for accurately evaluating whether a response aligns with the characters.
Furthermore, we transform the traditional subjective scoring approach into a novel bidirectional-evidence-finding task, which significantly improves the human agreement of LLM-as-Judge evaluations.

To develop a benchmark, the primary problem is the construction of a dataset. However, existing datasets fail to support semantically decoupled evaluation on EC and RC.
On the one hand, existing multimodal emotional dialogue datasets lack role-related data (such as role profiles and role images) and detailed semantic descriptions of multimodal data.
On the other hand, existing multimodal role-playing datasets mostly adopt synthetic methods, lacking in-depth integration of real dynamic emotions.
To address this issue, we construct a new dataset {\ProposedDatasetMERRY} based on the video materials from CPED\footnote{CPED, a large-scale Chinese Personalized and Emotional Dialogue dataset, which contains multi-turn multimodal dialogue data originating from TVs, and several labels including 13 types of emotions and 19 types of dialogue actions on utterance level, and gender, age, Big Five personality traits on role level.} \cite{chen2022cped}.

In conclusion, our main contributions can be summarized as follows:
\begin{enumerate}[itemsep=0pt, topsep=0pt, parsep=0pt, partopsep=0pt] 
	\item \textbf{We propose {\ProposedFrameworkMERRY}, a semantically decoupled evaluation framework}, addressing the limitation of existing end-to-end multimodal evaluations, where it is often ambiguous whether errors stem from the capabilities of MLLMs or the synthesis quality of talker modules. 
	\item \textbf{Fine-grained and Robust Metrics}: We introduce eight refined metrics for Emotional Consistency (EC) and Role Consistency (RC). Notably, we transform the traditional subjective RC scoring into a novel \enquote{bidirectional-evidence-finding} task, which significantly improves the reliability of LLM-as-Judge evaluations.
	\item \textbf{High-Quality Dataset}: Derived from real-world videos (CPED), we construct {\ProposedDatasetMERRY}, a multi-turn multimodal role-playing emotional interaction dialogue dataset, which includes role-related information and detailed semantic descriptions for facial expressions, body movements, and speech prompts, enabling precise, decoupled training and evaluation of MRPAs.  
	\item \textbf{Highlighted Findings}: {\Findings}
\end{enumerate}

\begin{table}[t]
	\centering
	\caption{Existing Role-Playing Benchmarks.} 
	\label{table:RPbenchs}
	\resizebox{\linewidth}{!}{
		\begin{tabular}{@{}ccccc@{}}
			\toprule
			& \begin{tabular}[c]{@{}c@{}}Semantically Decoupled\\MM Expression Eval\end{tabular} & EC           & RC           & \begin{tabular}[c]{@{}c@{}}Cross-modal\\ Eval\end{tabular} \\ \midrule
			\begin{tabular}[c]{@{}c@{}}CharacterEval \cite{tu-etal-2024-charactereval}\end{tabular}   & \ding{56}                                                                                  & \ding{56}           & \ding{52}          & \ding{56}                                                        \\
			\begin{tabular}[c]{@{}c@{}}CharacterBox \cite{wang2024characterboxevaluatingroleplayingcapabilities}\end{tabular}    & \ding{56}                                                                                  & \ding{56}           & \ding{52}          & \ding{56}                                                        \\
			\begin{tabular}[c]{@{}c@{}}InCharacter \cite{wang-etal-2024-incharacter}\end{tabular}     & \ding{56}                                                                                  & \ding{56}           & \ding{52}          & \ding{56}                                                        \\
			\begin{tabular}[c]{@{}c@{}}TimeChara \cite{ahn-etal-2024-timechara}\end{tabular}     & \ding{56}                                                                                  & \ding{56}           & \ding{52}          & \ding{56}                                                        \\
			\begin{tabular}[c]{@{}c@{}}MMRole \cite{dai2025mmrolecomprehensiveframeworkdeveloping}\end{tabular}          & \ding{56}                                                                                  & \ding{56}           & \ding{52}          & \ding{56}                                                        \\
			\begin{tabular}[c]{@{}c@{}}Beyond Dialogue \cite{yu-etal-2025-beyond}\end{tabular} & \ding{56}                                                                                  & \ding{52}          & \ding{52}          & \ding{56}                                                        \\
			\begin{tabular}[c]{@{}c@{}}EmoCharacter \cite{feng-etal-2025-emocharacter}\end{tabular}    & \ding{56}                                                                                  & \ding{52}          & \ding{56}           & \ding{56}                                                        \\
			\textbf{\ProposedFrameworkMERRY~(Ours)}   & \textbf{\ding{52}}                                                                        & \textbf{\ding{52}} & \textbf{\ding{52}} & \textbf{\ding{52}} \\  \bottomrule
		\end{tabular}
	}
\end{table}

\begin{table*}[ht]
	\caption{Comparison with Existing Dataset. a, v, and l refer audio, visual, and text respectively. MM Refers to MultiModal. EN and CN Refer to English and Chinese respectively.}
	\label{table:dataset_comparison}
	\centering
	\resizebox{\textwidth}{!}{
		\begin{tabular}{@{}cccccccc@{}}
			\toprule
			\textbf{Dataset} &
			\begin{tabular}[c]{@{}c@{}}\textbf{Modalities}\\\textbf{(per turn)}\end{tabular} &
			\textbf{Data Type} &
			\begin{tabular}[c]{@{}c@{}}\textbf{Role}\\\textbf{Profiles}\end{tabular} &
			\begin{tabular}[c]{@{}c@{}}\textbf{Emotion}\\\textbf{Labels}\end{tabular} &
			\begin{tabular}[c]{@{}c@{}}\textbf{Semantic Descriptions}\\\textbf{of MM data}\end{tabular} &
			\textbf{Language} &
			\textbf{Dialogue Source} \\ \midrule
			\multicolumn{8}{l}{\textit{Multimodal Emotional/Empathetic Dialogue Datasets}} \\ \midrule
			\begin{tabular}[c]{@{}c@{}}MELD \cite{poria-etal-2019-meld}\end{tabular} &
			(a, v, l) &
			MM dialogue &
			\ding{56} &
			\ding{52} &
			\ding{56} &
			EN &
			\enquote{Friends}   TV \\
			\begin{tabular}[c]{@{}c@{}}M3ED \cite{zhao-etal-2022-m3ed}\end{tabular} &
			(a, v, l) &
			MM dialogue &
			\ding{56} &
			\ding{52} &
			\ding{56} &
			CN &
			56   TVs \\
			\begin{tabular}[c]{@{}c@{}}CPED \cite{chen2022cped}\end{tabular} &
			(a, v, l) &
			MM dialogue &
			\ding{56} &
			\ding{52} &
			\ding{56} &
			CN &
			40   TVs \\
			\begin{tabular}[c]{@{}c@{}}MC-EIU \cite{liu2024emotionintentjointunderstanding}\end{tabular} &
			(a, v, l) &
			MM dialogue &
			\ding{56} &
			\ding{52} &
			\ding{56} &
			EN\&CN &
			\begin{tabular}[c]{@{}c@{}}3 English TVs\\ 4 Chinese TVs\end{tabular} \\
			\begin{tabular}[c]{@{}c@{}}MESC \cite{chu2024multimodalemotionalsupportconversation}\end{tabular} &
			(a, v, l) &
			MM dialogue &
			\ding{56} &
			\ding{52} &
			\ding{56} &
			EN &
			\enquote{In Treatment} TV \\
			\begin{tabular}[c]{@{}c@{}}AvaMERG \cite{zhang2025multimodalempatheticresponsegeneration}\end{tabular} &
			(a, v, l) &
			MM dialogue &
			\ding{56} &
			\ding{52} &
			\ding{56} &
			EN &
			\begin{tabular}[c]{@{}c@{}}Text: ED\cite{li2021knowledgebridgingempatheticdialogue}\\ MM: human acting\end{tabular} \\ \midrule
			\multicolumn{8}{l}{\textit{Role-Playing Datasets}} \\ \midrule
			\begin{tabular}[c]{@{}c@{}}Character-LLM \cite{shao-etal-2023-character}\end{tabular} &
			(\_, \_, l) &
			Pure Text dialogue &
			\ding{52} &
			\ding{56} &
			\ding{56} &
			EN &
			Synthesis \\
			\begin{tabular}[c]{@{}c@{}}CharacterGLM \cite{zhou-etal-2024-characterglm}\end{tabular} &
			(\_, \_, l) &
			Pure Text dialogue &
			\ding{52} &
			\ding{56} &
			\ding{56} &
			CN &
			\begin{tabular}[c]{@{}c@{}}Real-world, Literary,\\ and Synthesis\end{tabular} \\
			\begin{tabular}[c]{@{}c@{}}CharacterEval-RM \cite{tu-etal-2024-charactereval}\end{tabular} &
			(\_, \_, l) &
			Pure Text dialogue &
			\ding{52} &
			\ding{56} &
			\ding{56} &
			CN &
			Novels \\
			\begin{tabular}[c]{@{}c@{}}MMRole \cite{dai2025mmrolecomprehensiveframeworkdeveloping}\end{tabular} &
			(\_, \_, l) &
			MM based Text dialogue &
			\ding{52} &
			\ding{56} &
			\ding{56} &
			EN\&CN &
			Synthesis \\
			\begin{tabular}[c]{@{}c@{}}OmniCharacter-10K \cite{zhang2025omnicharacterimmersiveroleplayingagents}\end{tabular} &
			(a, \_, l) &
			MM dialogue &
			\ding{52} &
			\ding{56} &
			\ding{56} &
			EN &
			Synthesis \\ \midrule
			\textbf{\begin{tabular}[c]{@{}c@{}}\ProposedDatasetMERRY\\(Ours)\end{tabular}} &
			\textbf{(a, v, l)} &
			\textbf{MM dialogue} &
			\textbf{\ding{52}} &
			\textbf{\ding{52}} &
			\textbf{\ding{52}} &
			\textbf{CN} &
			\textbf{\begin{tabular}[c]{@{}c@{}}CPED \cite{chen2022cped}\end{tabular}} \\ \bottomrule
		\end{tabular}
	}
\end{table*}

\section{Related Work}
\subsection{Evaluation of Role-Playing Agents}
To evaluate the accuracy and vividness of Role-Playing Agents (RPAs), existing research generally focuses on three dimensions: knowledge \cite{tu-etal-2024-charactereval,ahn-etal-2024-timechara}, personality \cite{wang-etal-2024-incharacter,wang2024characterboxevaluatingroleplayingcapabilities,yu-etal-2025-beyond}, and emotional consistency \cite{feng-etal-2025-emocharacter}, where both knowledge and personality fall under the category of role consistency.
As shown in Table \ref{table:RPbenchs}, these evaluations of RPAs are based on the content of text responses, lacking assessments that semantically decouple multimodal emotional expressions.
Notably, dimensions regarding general capabilities, such as dialogue ability \cite{dai2025mmrolecomprehensiveframeworkdeveloping} and empathy \cite{tu-etal-2024-charactereval}, are beyond the scope of this paper.
Furthermore, existing automatic modality-synthesis metrics, including audio (e.g., WER, CER, SS) and video (e.g., CPBD, SSIM, $Sync_{cf}$) metrics, primarily focus on low-level perceptual or signal fidelity, but fail to evaluate the underlying semantic-level properties such as cross-modal consistency, emotional expressiveness, and role-style consistency, thereby still relying primarily on human judgment \cite{chen2025emoavatar,zhang2025multimodalempatheticresponsegeneration}.

\subsection{Multimodal Role-Playing Datasets}
Existing role-playing datasets can be divided into two aspects: 
\begin{enumerate*}[label=(\arabic*), itemsep=0pt, topsep=0pt, parsep=0pt, partopsep=0pt] 
	\item textual role-playing datasets \cite{shao-etal-2023-character, zhou-etal-2024-characterglm, wang-etal-2024-rolellm, wu-etal-2024-role, li2023chatharuhirevivinganimecharacter, chen-etal-2023-large};
	\item multimodal role-playing datasets \cite{dai2025mmrolecomprehensiveframeworkdeveloping, zhang2025omnicharacterimmersiveroleplayingagents}.
\end{enumerate*}
However, these datasets are primarily constructed by synthetic methods, lacking in-depth integration of real dynamic emotions. 
On the other hand, most early multimodal emotional \cite{chen2022cped, poria-etal-2019-meld, zhao-etal-2022-m3ed, liu2024emotionintentjointunderstanding} or empathetic \cite{zhang2025multimodalempatheticresponsegeneration, chu2024multimodalemotionalsupportconversation} dialogue datasets lack role-related data and detailed semantic descriptions of multimodal data. These datasets fail to support semantically decoupled evaluation on EC and RC, thereby motivating the construction of proposed dataset {\ProposedDatasetMERRY}. Detailed comparison can be found in Table~\ref{table:dataset_comparison}.

\begin{figure*}[ht]
	\centering
	\includegraphics[width=\textwidth]{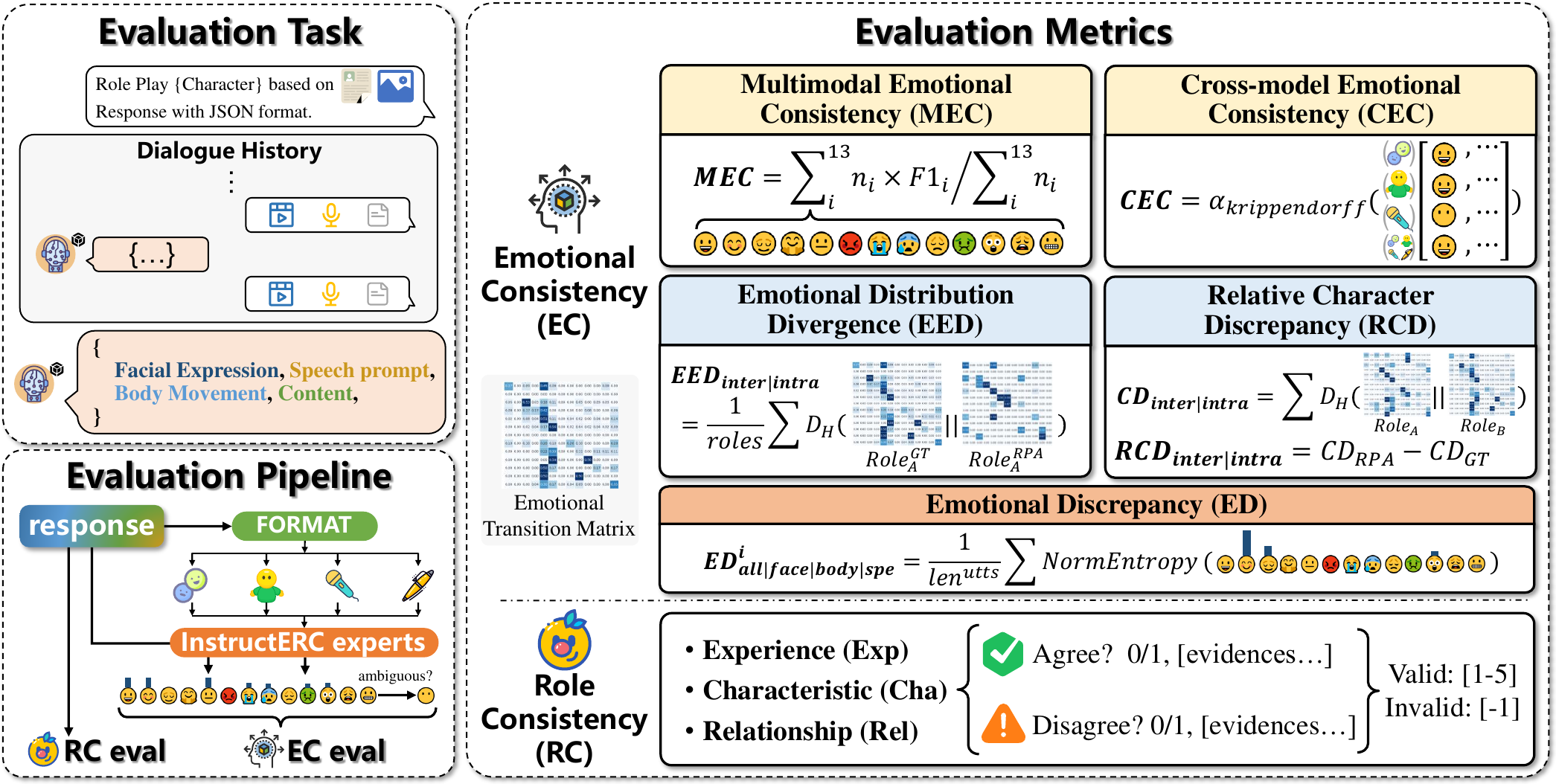} 
	\caption{Framework of {\ProposedFrameworkMERRY}. \textbf{Upper Left}: dialogue format of evaluation tasks.  \textbf{Lower Left}: evaluation pipeline.  \textbf{Right}: eight metrics for emotional and role consistencies.} 
	\label{fig:Overview_of_proposed_method}
\end{figure*}

\section{MERRY}

\subsection{Preliminaries}
\subsubsection{Evaluation Task Definition}
Given the role profile $P_C$ and role image $I_C$ (for self-localization in videos) of character $C$, RPAs step into shoes of this character and engage in multi-turn multimodal emotional dialogues with user named $N_u$, and previous information on current dialogue $Prev_D$:
\begin{align} 
    e_i &= \mathrm{RPAs}(Prof_c,I_C,N_u,Prev_D,D,u_i) \\
    D &= \{u_0,e_0,...,u_{i-1},e_{i-1}\}
\end{align}
where $D$ denotes a dialogue history sequence, $u_i= \{A_i,V_i,T_i\}$ is the $i$-th input from user, including audio $A_i$, video $V_i$, and content $T_i$, and $e_i=\{f_i,b_i,s_i,c_i\}$ as the $i$-th turn response from RPAs, including textual response content $c_i$, facial expression $f_i$, body movement $b_i$, and speech prompt $s_i$ (namely speech intonation). Notably $e_i$ can be formatted as JSON, which is most widely used and trained in various base models. $u_0$ and $e_0$ are the special input and output cases when $i = 0$, respectively.



\subsubsection{Emotion Transition in Dialogue}
\label{sec:Emotion Transition in Dialogue}
Different characters exhibit unique emotional changes during conversations, which can be captured by emotional transition matrices $M$ \cite{feng-etal-2025-emocharacter}.
Furthermore, a response often comprises multiple utterances separated by punctuation, which may convey different emotions. This suggests that emotional shifts can occur not only inter-turn, but also intra-turn.
Therefore, we further decompose $M$ into $M_{intra}$ and $M_{inter}$, where the former counts the emotional transitions during each character's expression process, and the latter counts the emotional changes of the character towards different inputs.
Specifically, given a sample $i$ with dialogue history $D = \{u_0,e_0,...,u_{i-1},e_{i-1}\}$, user input $u_i$, and target or prediction from system $e_i$, the definition of inter-turn emotional transition matrix remains as \cite{feng-etal-2025-emocharacter}, whereas our definition of intra-turn one captures emotion transitions within a single response:
\begin{align}
	M_{inter}&=\sum_{i} TranPair(e_{i-1}, e_i) \\
    M_{intra}&=\sum_{i} \sum_{u}^{len^{utts}_i} TranPair(e^i_{u-1}, e^i_u)
\end{align}
where $len^{utts}_i$ denotes the number of utterances obtained by splitting textual response content $c_i$ with Chinese punctuation, which also represents the utterance length of the entire multimodal emotional expression $e_i$.


\subsection{Evaluation Pipeline}
As shown in Figure~\ref{fig:Overview_of_proposed_method} (Lower Left), RC evaluation can directly assess the predicted responses, whereas EC evaluation, requires an additional formatting and emotion recognition pipeline before evaluation. 
In this way, the output format is unified despite a broken response prediction, thereby avoiding format bias. Meanwhile, using emotion recognition results from the same group of experts, rather than RPA-generated emotions around response \cite{feng-etal-2025-emocharacter}, further mitigates model bias.
Specifically, the FORMAT step first attempts to validate the raw output using Pydantic model. If it fails, which is probably due to format issues such as missed quotation marks, 
Doubao-seed-1-6\footnote{\url{https://www.doubao.com/}} is invoked to reformat the $resp^{raw}_i$ into JSON-formatted output $resp^{fmt}_i=\{f_i,b_i,s_i,c_i\}$ without altering its semantics.
When it comes to ERC (Emotional Recognition in Conversation) step, inspired by InstructERC \cite{lei2023instructerc}, 
five experts are asked to conduct emotion recognition twice for each response $i$ with $len^{utts}_i$ utterances: 
\begin{align} 
	r^{ERC}_{i,j} &= ERC_{exp_j}(len^{utts}_i,resp^{fmt}_i) 
\end{align}
where $j$ denotes expert index.
The result $r^{ERC}_{i,j} = \{emos_{f},emos_{b},emos_{s},emos_{fusion}\}$ contains four recognized emotion lists, for facial expression, body movement, speech prompt, and all-modality fusion, respectively. Each emotion list have the same length as $len^{utts}_i$, for instance, $emos_{fusion}=\{emo_u\}_{u=1}^{len^{utts}_i}$.
Generally, given a formatted response $resp^{fmt}_i$ with $len^{utts}_i$ utterances, each modality utterance is associated with ten recognized emotion labels from different experts, forming an emotional distribution $EmoD_{u}$, and the emotion label $emo$ with probability $prob^{emo}_{utt} \ge \tau$ is selected as the final label of the utterance, otherwise it will be marked as \enquote{\textit{ambiguous}} and excluded from the subsequent EC evaluation.
Specifically, in this paper, $\tau$ is set as 0.7, which means a dominant recognition result agreed by more than two-thirds of the votes. 
Besides, activated ERC experts are latest models from different series, including Doubao-seed-1-6, GPT-5.2\footnote{\url{https://openai.com/api/}}, Gemini-3-flash\footnote{\url{https://gemini.google.com/}}, 
Claude-sonnet-4.5\footnote{\url{https://claude.ai/}}, Deepseek-V3.2\footnote{\url{https://deepseek.ai/}}. 

\subsection{Evaluation Metrics}
As shown in Figure~\ref{fig:Overview_of_proposed_method} (Right), we design five metrics for Emotional Consistency and three for Role Consistency.

\subsubsection{Emotional Consistency (EC)}
\textbf{EC} originally consists of micro Emotional Consistency(EC-micro), Emotional Distribution Divergence(EDD), and Relative Character Discrepancy(RCD), in the context of text conversations \cite{feng-etal-2025-emocharacter}, quantifies how well RPAs can simulate the real and dynamic emotions of designated roles.

In order to measure how precisely and accurately RPAs simulate the real emotions of the designated roles within multimodal dialogues,
\textit{EC-micro} should further integrate all multimodal expressions into account, thereby motivating the proposed \textbf{Multimodal Emotional Consistency (MEC)}:
\begin{align}
	MEC_{lower|upper}=\frac{\sum_{x=1}^{13} n_x \times F1_x }{\sum_{x=1}^{13} n_x} 
\end{align}
where $n_1, n_2, ..., n_{13}$ and $F1_1, F1_2, ..., F1_{13}$ denotes the number of testset samples and their respective F1 scores among the 13 different emotion categories (same as CPED).
Confusion matrix is as shown in Table~\ref{tab:ec_f1_map_rule},
where $set()$ function is applied because multiple identical emotions might appear in a single response.
Similarly, $MEC_{lower}$ denotes matching the exact emotion label, while $MEC_{upper}$ matches the emotional tendency instead.
Both two MEC scores range from 0 to 1, with higher values indicating better performance.
Notably, we selectively omit the original middle value based on Valence-Arousal-Dominance (VAD) emotional vector, since a fixed VAD vector introduces bias for different languages, dialogue context, and roles.

\textbf{Emotional Distribution Divergence(EDD)} originally measures the realism of dynamic emotional transitions in portrayed characters by comparing them with those observed in real ones,
whereas \textbf{Relative Character Discrepancy(RCD)} evaluates whether role differences, as reflected by emotional transition matrices, are sufficiently distinguishable yet not overly exaggerated.
Furthermore, as discussed in Section~\ref{sec:Emotion Transition in Dialogue}, emotional transition matrices are decomposed into $M_{intra}$ and $M_{inter}$.
Therefore, EDD and RCD is also decomposed into $EDD_{inter}$ and $EDD_{inter}$, $RCD_{inter}$ and $RCD_{inter}$, respectively:
\begin{align}
	EED_{inter|intra}&=\frac{\sum_{k}^{K} D_H(M_{inter|intra}^{GT_k}||M_{inter|intra}^{RPA_k})}{K} \\
	CD_{inter|intra}=&\frac{\sum_{role_A,role_B}^{\binom{2}{K}} D_H(M_{inter|intra}^{role_A}||M_{inter|intra}^{role_B})}{\binom{2}{K}}\\
	RCD_{inter|intra}&=CD_{inter|intra}^{RPA} - CD_{inter|intra}^{GT}
\end{align}
where $K$ is the number of roles and $CD_{inter|intra}^{GT}$ is a constant value across different RPAs.
Notably, the original unbounded and asymmetric KL distance is replaced as the bounded and symmetric Hellinger distance \cite{Hellinger+1909+210+271} (ranging from 0 to 1):
\begin{align}
	D_H(P||Q)=\frac{1}{\sqrt{2}} \sqrt{\sum_i {(\sqrt{p_i}-\sqrt{q_i})}^2}
\end{align}
Therefore, EED is bounded from 0 to 1, with lower values indicating better performance; whereas RCD is bounded from -1 to 1, with values closer to 0 denoting better alignment with real roles.
Specifically, if RCD $>$ 0 or increase, it suggests that the model overly emphasize the character differences.
By contrast, this implies that the emotional transitions between roles are too similar, failing to provide sufficient differentiation.

\begin{table}[]
	\caption{Confusion matrix. $emo_x$ denotes the x-th class emotion. \textit{GT} and \textit{PD} refers to $emos_{gt}$ and $emos_{fusion}$, respectively. \enquote{set(list)} function is used to convert the list into unique elements.}
	\resizebox{\linewidth}{!}{
		\begin{tabular}{@{}ccc@{}}
			\toprule
			& $emo_x$ in set(PD) & $emo_x$ not in set(PD) \\ \midrule
			$emo_x$ in set(GT)     & TP             & FN                 \\
			$emo_x$ not in set(GT) & FP             & TN                 \\ \bottomrule
		\end{tabular}
	}
	\centering
	\label{tab:ec_f1_map_rule}
\end{table}

When there are multimodal emotional expressions, EC should further measure the degree of emotional consistency across different modalities, namely \textbf{Cross-model Emotional Consistency (CEC)}.
Otherwise, inconsistent emotions across modalities will cause a severe sense of dissonance, undermining the immersive experience.
During assessment, Krippendorff's alpha \cite{Krippendorff2011ComputingKA} is applied as metric, since it supports multiple raters and invalid values:
\begin{align}
	CEC_{lower|upper}=\alpha_{krippendorff}(\begin{bmatrix}
		emos_{f} &  ... \\
		emos_{b} &  ... \\
		emos_{s} &  ... \\
		emos_{fusion} &  ...
	\end{bmatrix})
\end{align}
where the four modalities are considered as different raters. Similarly, $CEC_{lower}$ and $CEC_{upper}$ matches exact emotion labels and their tendencies, respectively.

Finally, we introduce \textbf{Emotional Discrepancy (ED)} for quantifying how much ambiguity is in the emotion recognition experts, reflecting the precision of emotions conveyed by multimodal expressions to a certain extent:
\begin{align}
	ED_{fus|f|b|s}&=\\
	&\frac{\sum_i \sum_u NormEntropy(EmoD_{u,i}^{fus|f|b|s})}{\sum_i len^{utts}_i}
\end{align}
where $EmoD_{u,i}^{fus|f|b|s}$ denotes the emotional distribution of utterance $u$ in sample $i$, calculated on different modalities.

\begin{table}[]
	\centering
	\caption{Role consistency mapping table.}
	\resizebox{\linewidth}{!}{
		\begin{tabular}{@{}ccccccc@{}}
			\toprule
			AgreeFlag                                                            & 0                & 1                & 1              & 1 & 1           & 0                \\
			DisagreeFlag                                                         & 0                & 0                & 1              & 1 & 1           & 1                \\
			\begin{tabular}[c]{@{}c@{}}evidences\\ (agree vs. disagree)\end{tabular} & \textbackslash{} & \textbackslash{} & \textgreater{} & = & \textless{} & \textbackslash{} \\ \midrule
			FinalScore                                                           & dropped               & 5                & 4              & 3 & 2           & 1                \\ \bottomrule
		\end{tabular}
	}
	\label{tab:rc_map_rule}
\end{table}

\subsubsection{Role Consistency (RC)}
LLMScore \cite{gu2025surveyllmasajudge} is primarily used when evaluating RC. 
To improve the consistency between human evaluators and LLMs, we propose to turn the traditional subjective scoring into bidirectional-evidence-finding approach. The final score mapping rule is as exhibited in Table~\ref{tab:rc_map_rule}. 
In this way, we improve the consistency and robustness while maintaining the granularity.
Human agreement experimental results in Table~\ref{tab:human_agreement_metric} further demonstrate the effectiveness of proposed approach.
During evaluation, we activate two evaluators, including Doubao-seed-1-6 and GPT-5.2. The final score is the average score of two experts.
\begin{figure}[t]
	\centering
	\includegraphics[width=\columnwidth]{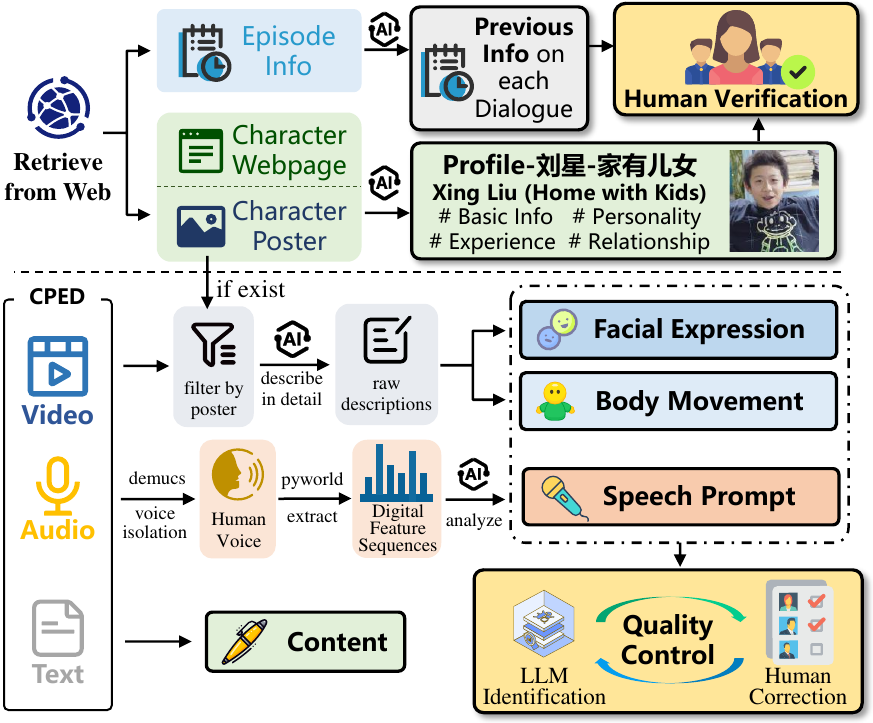} 
	\caption{\ProposedDatasetMERRY~Construction Pipeline} 
	\label{fig:data_construction_pipeline}
\end{figure}

\subsection{Construction of {\ProposedDatasetMERRY}}
As shown in Figure~\ref{fig:data_construction_pipeline}, we construct {\ProposedDatasetMERRY}, a multi-turn multimodal role-playing emotional interaction dialogue dataset, based on CPED \cite{chen2022cped}.
\textbf{Role-related information:}
For role profiles and images, we first retrieve role webpage and posters from BaiduBaike\footnote{\url{https://baike.baidu.com/}, a Chinese Encyclopedia}.
After filtering out missed roles and corresponding dialogues in dataset, we use Deepseek-V3 to summarize into profiles containing four primary parts: basic information, characteristics, experiences, and interpersonal relationships.
For previous information on each dialogue, we retrieve episode information from TVMAO\footnote{\url{https://www.tvmao.com/}, a public website including outlines for various Chinese TV episodes}, and use Doubao-seed-1-6 to generate previous information for each dialogue.
Finally we recruit a team of 20 volunteer students from school to check and correct the summarized profiles and role images based on corresponding webpages, and the previous information based on webpages, dialogues, and their episode information.

\textbf{Multimodal Detailed Semantic Descriptions:}
We propose using detailed semantic descriptions for semantically decoupled training and evaluation, including facial expressions, body movements, and speech prompts (namely intonations), each emotionally distinguishable.
First, we generate coarse-grain semantic descriptions.
For audio modality, we follow current studies \cite{Shimizu-etal-PromptTTS++,liu-et-al-2024-NCSSD} to obtain normalized numerical sequences of \textit{speaking speed}, \textit{pitches} and \textit{volumes} from audios. 
We then use Deepseek-V3 to analyze those numerical feature sequences and generate speech prompt.
For video modality, we directly use GPT4o-mini \cite{openai2024gpt4ocard} to describe facial expressions and body movements based on role images and videos (converted into frames).

Second, we further conduct human-in-the-loop quality control, leveraging the power of cooperation between human and LLMs.
Specifically, we first instruct Doubao-seed-1-6 to separately identify issued descriptions based on video (given frames) and audio (given numerical sequences), and providing suggestions for correction.
Issued descriptions are defined as mismatching (1) videos, (2) emotion labels.
Subsequently, we instruct a team of 20 volunteers to verify these issued descriptions with suggestions.
After two rounds of searching and correction, we acquire a high-quality dataset in the end.

\textbf{Assemble and Statistics:}
In this step, we assemble all components above and retain those dialogues with two speakers only.
Finally we divide the dataset into a train set and a test set based on TV series to strictly avoid data leakage.
The statistics are as shown in Tables~\ref{tab:statistics_splits} and \ref{tab:statistics}.

\begin{table}[ht]
	\centering
	\caption{Statistics of train and test splits of \ProposedDatasetMERRY.}
	\label{tab:statistics_splits}
	\resizebox{0.8\columnwidth}{!}{
		\begin{tabular}{@{}cccc@{}}
			\toprule
			Split & \# Samples & \# Roles & \# Dialogues \\ \midrule
			Train & 23817      & 239      & 3577 \\
			Test  & 1690       & 18       & 401  \\ \bottomrule
		\end{tabular}
	}
\end{table}

\begin{table}[ht]
	\caption{Statistics of whole proposed dataset \ProposedDatasetMERRY.}
	\centering
	\label{tab:statistics}
	\resizebox{0.95\columnwidth}{!}{
		\begin{tabular}{l|c}
			\toprule
			\#samples                             & 25507  \\
			\#dialogues                           & 3978   \\
			Avg. tokens of previous information   & 35.33 \\
			\#roles                               & 257    \\ \midrule
			Avg. of \#utterances per role         & 114.73 \\
			Avg. of \#turns per dialogue          & 7.41 \\
			Avg. durations of video (seconds)     & 6.00   \\ \midrule
			Avg. tokens of facial expression  & 145.76 \\
			Avg. tokens of body movement      & 142.52 \\
			Avg. tokens of speech prompt      & 224.58  \\
			Avg. tokens of content            & 34.38  \\
			Avg. tokens of role profile           & 518.36 \\ \bottomrule
		\end{tabular}
	}
\end{table}

\section{Experiments}

\subsection{Experimental Setup}
\label{experiment_settings}
In general, we conduct two set of comparison experiments based on proposed evaluation framework {\ProposedFrameworkMERRY}, including training-free and training methods.

For training-free role-playing methods, we evaluate three commonly used close-source MLLMs, including Doubao-1.5v (full name is Doubao-1.5-thinking-vision-pro) \cite{guo2025seed15vltechnicalreport},
Gemini-2.5-pro \cite{comanici2025gemini25pushingfrontier},
and GPT-5-chat \cite{singh2025openaigpt5card},
and two existing open-source base models, Qwen2.5-Omni(7B) \cite{Qwen2.5-Omni} and MiniCPM-o-2.6(8B) \cite{yao2024minicpm}.
Notably, we selectively 
During inference, we use the same generation config for all baselines, specifying a temperature of 0.7, top-p of 0.95, while other settings remain as the default configurations of each base model. 
Notably, since Doubao-1.5v and GPT-5-chat are confined to processing vision-only videos, and fail to directly chat on audio files, we replace audio input files of the dialogue context by their corresponding speech prompts for these models.

For training methods, we fine-tune a unified base model MiniCPM-o-2.6(8B) with LoRA \cite{hu2021loralowrankadaptationlarge} for fair comparison, using 1*A800 80GB GPU on the different training sets. 
Uniformly, we set $lora\_rank$ as 16, $lora\_alpha$ as 32, $lora\_dropout$ as 0.05, and adopt a learning rate equals to 1e-5 with cosine scheduler type and warm-up ratio as 0.1.
Notably we activate a validation set randomly sampled from train set with a ratio of 0.01.

\begin{table*}[]
	\centering
	\caption{Performance results of base models with training-free role-playing methods on the entire testset. Best results are in \textbf{bold}, and the second best result are \underline{underlined}. Type \textit{Prev}, \textit{Prof}, \textit{All} denotes that the model is equipped with previous information, profile, both of them, respectively. While \textit{None} denotes that the model are asked to play a role without profile and previous information. $\dagger$ means the audio files are replaced as speech prompts for these models.}
	\label{tab:agent results}
	\resizebox{\textwidth}{!}{%
		\begin{tabular}{ccccccccccccccccc}
			\toprule
			\multirow{2}{*}{\textbf{Models}} &
			\multirow{2}{*}{\textbf{Type}} &
			\multicolumn{2}{c}{\textbf{MEC $\uparrow$}} &
			\multicolumn{2}{c}{\textbf{CEC $\uparrow$}} &
			\multicolumn{2}{c}{\textbf{EDD $\downarrow$}} &
			\multicolumn{2}{c}{\textbf{RCD $\leftrightarrow$}} &
			\multicolumn{4}{c}{\textbf{ED $\downarrow$}} &
			\multicolumn{3}{c}{\textbf{RC $\uparrow$}} \\ \cmidrule(l){3-4} \cmidrule(l){5-6} \cmidrule(l){7-8} \cmidrule(l){9-10} \cmidrule(l){11-14} \cmidrule(l){15-17} 
			&
			&
			\textbf{lower} &
			\textbf{upper} &
			\textbf{lower} &
			\textbf{upper} &
			\textbf{intra} &
			\textbf{inter} &
			\textbf{intra} &
			\textbf{inter} &
			\textbf{all} &
			\textbf{spe} &
			\textbf{fac} &
			\textbf{bod} &
			\textbf{Exp} &
			\textbf{Cha} &
			\textbf{Rel} \\ \midrule
			\multicolumn{17}{c}{\textit{Close-source MLLMs}} \\ \midrule
			\multirow{4}{*}{Doubao-1.5v $\dagger$} &
			All &
			\textbf{0.243} &
			\textbf{0.546} &
			\textbf{0.658} &
			\underline{0.745} &
			\textbf{0.410} &
			0.455 &
			-0.033 &
			\underline{-0.006} &
			0.648 &
			0.634 &
			0.632 &
			0.615 &
			\underline{4.891} &
			4.720 &
			\underline{4.877} \\
			&
			Prof &
			\underline{0.232} &
			0.489 &
			0.592 &
			\underline{0.745} &
			0.414 &
			\textbf{0.449} &
			\textbf{-0.017} &
			0.011 &
			\underline{0.610} &
			\underline{0.577} &
			\underline{0.596} &
			\underline{0.587} &
			4.634 &
			\textbf{4.805} &
			4.800 \\
			&
			Prev &
			\textbf{0.243} &
			\underline{0.524} &
			0.615 &
			\textbf{0.755} &
			0.425 &
			0.456 &
			-0.030 &
			\textbf{0.003} &
			\textbf{0.598} &
			\textbf{0.563} &
			\textbf{0.586} &
			\textbf{0.573} &
			\textbf{4.909} &
			\underline{4.804} &
			\textbf{4.923} \\
			&
			None &
			0.230 &
			0.504 &
			\underline{0.637} &
			0.727 &
			\underline{0.412} &
			\underline{0.454} &
			\underline{-0.026} &
			\underline{-0.006} &
			0.661 &
			0.636 &
			0.645 &
			0.644 &
			4.439 &
			4.527 &
			4.757 \\ \midrule
			\multirow{4}{*}{GPT-5-chat $\dagger$} &
			All &
			\underline{0.251} &
			\textbf{0.559} &
			\textbf{0.642} &
			0.731 &
			\textbf{0.404} &
			\underline{0.449} &
			\textbf{-0.003} &
			0.016 &
			\underline{0.649} &
			0.620 &
			0.644 &
			0.613 &
			\underline{4.862} &
			4.742 &
			\underline{4.887} \\
			&
			Prof &
			0.250 &
			0.521 &
			0.568 &
			0.707 &
			\underline{0.431} &
			0.461 &
			0.019 &
			0.022 &
			0.672 &
			0.619 &
			0.668 &
			\underline{0.593} &
			4.648 &
			\textbf{4.865} &
			4.846 \\
			&
			Prev &
			\textbf{0.257} &
			\underline{0.551} &
			0.614 &
			\underline{0.748} &
			0.443 &
			0.451 &
			0.029 &
			\underline{0.007} &
			\textbf{0.608} &
			\textbf{0.560} &
			\textbf{0.607} &
			\textbf{0.568} &
			\textbf{4.890} &
			\underline{4.849} &
			\textbf{4.917} \\
			&
			None &
			0.235 &
			0.533 &
			\underline{0.634} &
			\textbf{0.752} &
			0.434 &
			\textbf{0.433} &
			0.014 &
			\textbf{0.004} &
			\underline{0.649} &
			\underline{0.595} &
			\underline{0.643} &
			0.598 &
			4.378 &
			4.544 &
			4.691 \\ \midrule
			\multirow{4}{*}{Gemini-2.5-pro} &
			All &
			\textbf{0.279} &
			\textbf{0.608} &
			\textbf{0.720} &
			\underline{0.793} &
			\underline{0.393} &
			0.456 &
			\textbf{0.013} &
			\underline{0.019} &
			\underline{0.611} &
			0.585 &
			0.597 &
			0.548 &
			\underline{4.857} &
			4.629 &
			\underline{4.829} \\
			&
			Prof &
			0.258 &
			0.579 &
			0.679 &
			\textbf{0.797} &
			\textbf{0.389} &
			\underline{0.452} &
			0.026 &
			0.027 &
			\textbf{0.576} &
			\textbf{0.550} &
			\textbf{0.545} &
			\textbf{0.489} &
			4.681 &
			\textbf{4.760} &
			4.778 \\
			&
			Prev &
			\underline{0.265} &
			\underline{0.588} &
			0.675 &
			0.776 &
			0.437 &
			\textbf{0.432} &
			0.018 &
			\underline{0.019} &
			\textbf{0.576} &
			\underline{0.551} &
			\underline{0.555} &
			\underline{0.507} &
			\textbf{4.923} &
			\underline{4.740} &
			\textbf{4.896} \\
			&
			None &
			0.248 &
			\underline{0.588} &
			\underline{0.682} &
			0.761 &
			0.416 &
			0.458 &
			\underline{0.017} &
			\textbf{0.011} &
			0.635 &
			0.599 &
			0.616 &
			0.575 &
			4.439 &
			4.440 &
			4.611 \\ \midrule
			\multicolumn{17}{c}{\textit{Open-source MLLMs}} \\ \midrule
			\multirow{4}{*}{Qwen2.5-Omni(7B)} &
			All &
			\textbf{0.241} &
			\textbf{0.528} &
			\textbf{0.582} &
			0.682 &
			\textbf{0.401} & 
			0.421 &
			\textbf{-0.006} &
			\underline{-0.008} &
			0.620 &
			0.468 &
			0.447 &
			0.478 &
			\textbf{4.238} &
			4.175 &
			\textbf{4.389} \\
			&
			Prof &
			0.231 &
			0.499 &
			0.571 &
			\textbf{0.705} &
			0.404 &
			\textbf{0.415} &
			-0.034 &
			0.013 &
			\underline{0.546} &
			\underline{0.373} &
			\underline{0.388} &
			\underline{0.413} &
			3.824 &
			\textbf{4.178} &
			4.184 \\
			&
			Prev &
			\underline{0.237} &
			\underline{0.517} &
			\underline{0.576} &
			\underline{0.698} &
			0.409 &
			0.429 &
			\underline{-0.015} &
			\textbf{0.005} &
			\textbf{0.530} &
			\textbf{0.353} &
			\textbf{0.347} &
			\textbf{0.398} &
			\underline{4.045} &
			\underline{4.176} &
			\underline{4.348} \\
			&
			None &
			0.206 &
			0.470 &
			0.571 &
			0.683 &
			\underline{0.403} & 
			\underline{0.418} &
			-0.028 &
			-0.017 &
			0.618 &
			0.495 &
			0.451 &
			0.499 &
			3.683 &
			3.867 &
			4.114 \\ \midrule
			\multirow{4}{*}{MiniCPM-o-2.6(8B)} &
			All &
			\textbf{0.264} &
			\textbf{0.536} &
			0.586 &
			0.652 &
			0.381 &
			0.398 &
			\textbf{-0.002} &
			\underline{-0.017} &
			0.612 &
			0.537 &
			0.482 &
			0.471 &
			\textbf{4.026} &
			\textbf{4.117} &
			\textbf{4.170} \\
			&
			Prof &
			\underline{0.260} &
			0.524 &
			0.600 &
			\underline{0.686} &
			\textbf{0.356} &
			\underline{0.396} &
			\textbf{0.002} &
			\textbf{-0.011} &
			\underline{0.542} &
			\underline{0.444} &
			\underline{0.420} &
			\underline{0.364} &
			3.426 &
			4.017 &
			3.948 \\
			&
			Prev &
			0.258 &
			\underline{0.526} &
			\textbf{0.632} &
			\textbf{0.708} &
			\underline{0.368} &
			\textbf{0.386} &
			-0.049 &
			-0.025 &
			\textbf{0.500} &
			\textbf{0.398} &
			\textbf{0.384} &
			\textbf{0.368} &
			\underline{3.708} &
			\underline{4.037} &
			\underline{3.982} \\
			&
			None &
			0.243 &
			0.516 &
			\underline{0.612} &
			0.674 &
			0.369 &
			0.411 &
			\underline{-0.032} &
			-0.035 &
			0.584 &
			0.474 &
			0.432 &
			0.443 &
			3.220 &
			3.739 &
			3.708 \\ \bottomrule
		\end{tabular}%
	}
\end{table*}

\begin{table*}[]
	\centering
	\caption{Performance results of different training methods on the entire testset, with MiniCPM-o-2.6(8B) as the unified base model for fair comparison. Best results are in \textbf{bold}, and the second best result are \underline{underlined}.}
	\label{tab:train results}
	\resizebox{\textwidth}{!}{%
		\begin{tabular}{@{}ccccccccccccccccc@{}}
			\toprule
			\multirow{2}{*}{\textbf{Trainset}} &
			\multirow{2}{*}{\textbf{Method}} &
			\multicolumn{2}{c}{\textbf{MEC $\uparrow$}} &
			\multicolumn{2}{c}{\textbf{CEC $\uparrow$}} &
			\multicolumn{2}{c}{\textbf{EDD $\downarrow$}} &
			\multicolumn{2}{c}{\textbf{RCD $\leftrightarrow$}} &
			\multicolumn{4}{c}{\textbf{ED $\downarrow$}} &
			\multicolumn{3}{c}{\textbf{RC $\uparrow$}} \\ 			\cmidrule(l){3-4} \cmidrule(l){5-6} \cmidrule(l){7-8} \cmidrule(l){9-10} \cmidrule(l){11-14} \cmidrule(l){15-17} 
			&
			&
			\textbf{lower} &
			\textbf{upper} &
			\textbf{lower} &
			\textbf{upper} &
			\textbf{intra} &
			\textbf{inter} &
			\textbf{intra} &
			\textbf{inter} &
			\textbf{all} &
			\textbf{spe} &
			\textbf{fac} &
			\textbf{bod} &
			\textbf{Exp} &
			\textbf{Cha} &
			\textbf{Rel} \\ \midrule
			None &
			None &
			0.264 &
			\underline{0.536} &
			\underline{0.586} &
			\underline{0.652} &
			0.381 &
			\textbf{0.398} &
			\textbf{-0.002} &
			\underline{-0.017} &
			0.612 &
			0.537 &
			0.482 &
			0.471 &
			4.026 &
			4.117 &
			4.170 \\
			OmniCharacter &
			SFT &
			0.256 &
			0.534 &
			0.491 &
			0.514 &
			\textbf{0.365} &
			0.416 &
			\underline{-0.005} &
			\textbf{-0.001} &
			0.683 &
			0.451 &
			0.553 &
			0.471 &
			4.205 &
			\textbf{4.276} &
			\textbf{4.472} \\
			MMRole &
			SFT &
			\underline{0.275} & 
			0.517 &
			0.550 &
			0.586 &
			0.388 &
			\underline{0.402} &
			0.016 &
			-0.061 &
			\underline{0.539} &
			\underline{0.204} &
			\underline{0.167} &
			\underline{0.167} &
			\underline{4.267} &
			\underline{4.274} &
			\underline{4.417} \\
			MERRY (Ours) &
			SFT &
			\textbf{0.291} & 
			\textbf{0.541} &
			\textbf{0.934} &
			\textbf{0.962} &
			\underline{0.379} &
			0.412 &
			-0.024 &
			-0.021 &
			\textbf{0.257} &
			\textbf{0.117} &
			\textbf{0.106} &
			\textbf{0.152} &
			\textbf{4.357} &
			4.167 & 
			4.305 \\ \bottomrule
		\end{tabular}%
	}
\end{table*}

\subsection{Main Results}
Table~\ref{tab:agent results} reports the evaluation results of {\ProposedFrameworkMERRY} on different base models with four different training-free role-playing types, including \textit{All}, \textit{Prof} (profile only), \textit{Prev} (previous information only), and \textit{None}. We observe that:

(1) \textbf{Across most EC metrics, omni-modal foundation models consistently outperform vision-only counterparts}. Among closed-source models, \textit{Gemini-2.5-pro} is a multimodal foundation model, whereas \textit{Doubao-1.5v} and \textit{GPT-5-chat} are vision-only. Under the same type condition, regardless of which type, \textit{Gemini-2.5-pro} surpasses the other two vision-only models on all sub-metrics of $MEC$, $CEC$, and $ED$, except for $ED_{spe}$ when $type=None$. Notably, the open-source omni-modal foundation model \textit{MiniCPM-o-2.6(8B)} achieves an $MEC_{lower}$ score of 0.264, slightly exceeding those of the two closed-source vision-only models (0.243 and 0.251). This indicates that satisfying omni-modal input conditions is crucial for accurately understanding and consistently expressing characters' emotions. 

(2) For RC, all closed-source foundation models exhibit a consistent pattern. Specifically, for the $Exp$ and $Rel$ metrics, \textit{Prev} achieves the best performance, followed by \textit{All}, whereas for the $Cha$ metric, \textit{Prof} performs best, with \textit{Prev} as the second best. This pattern suggests that \textbf{\textit{Prev} provides important information for grounding potentially time-varying knowledge, while \textit{Prof} appears particularly beneficial for enabling RPAs to exhibit the intended character traits}.
However, when both types of information are provided simultaneously, namely \textit{All}, performance slightly decreases, which may indicate that longer contextual inputs introduce additional constraints that limit the models' flexibility in integrating information effectively, leading to a $1+1<2$ effect. By contrast, this pattern is reversed for open-source models. Type \textit{All} almost always yields the best performance (except for the $Cha$ metric of \textit{Qwen2.5-Omni(7B)}), exhibiting a $1+1>2$ effect.
This suggests that \textbf{long-context input may act as a constraint for stronger models, while serving as complementary information for weaker ones, revealing the limitations of current simple prompting methods}.


(3) For the $EDD_{intra}$ and $RCD_{intra}$ metrics, \textit{All} or \textit{Prof} consistently achieves the best performance across all tested foundation models. This is because intra-turn dynamic emotional transitions are often highly correlated with character personality traits. Therefore \textbf{providing profiles helps models express emotions more accurately and role-distinctively}, leading to higher $EDD_{intra}$ and $RCD_{intra}$, respectively.
In contrast, inter-turn emotional transitions additionally depend on varying inputs and dialogue history, and thus do not exhibit a clear association with any specific role-playing type. Consequently, some models even achieve optimal performance under $type=None$, such as \textit{GPT-5-chat}.


\begin{figure*}[htbp]
	\centering
	\includegraphics[width=\textwidth]{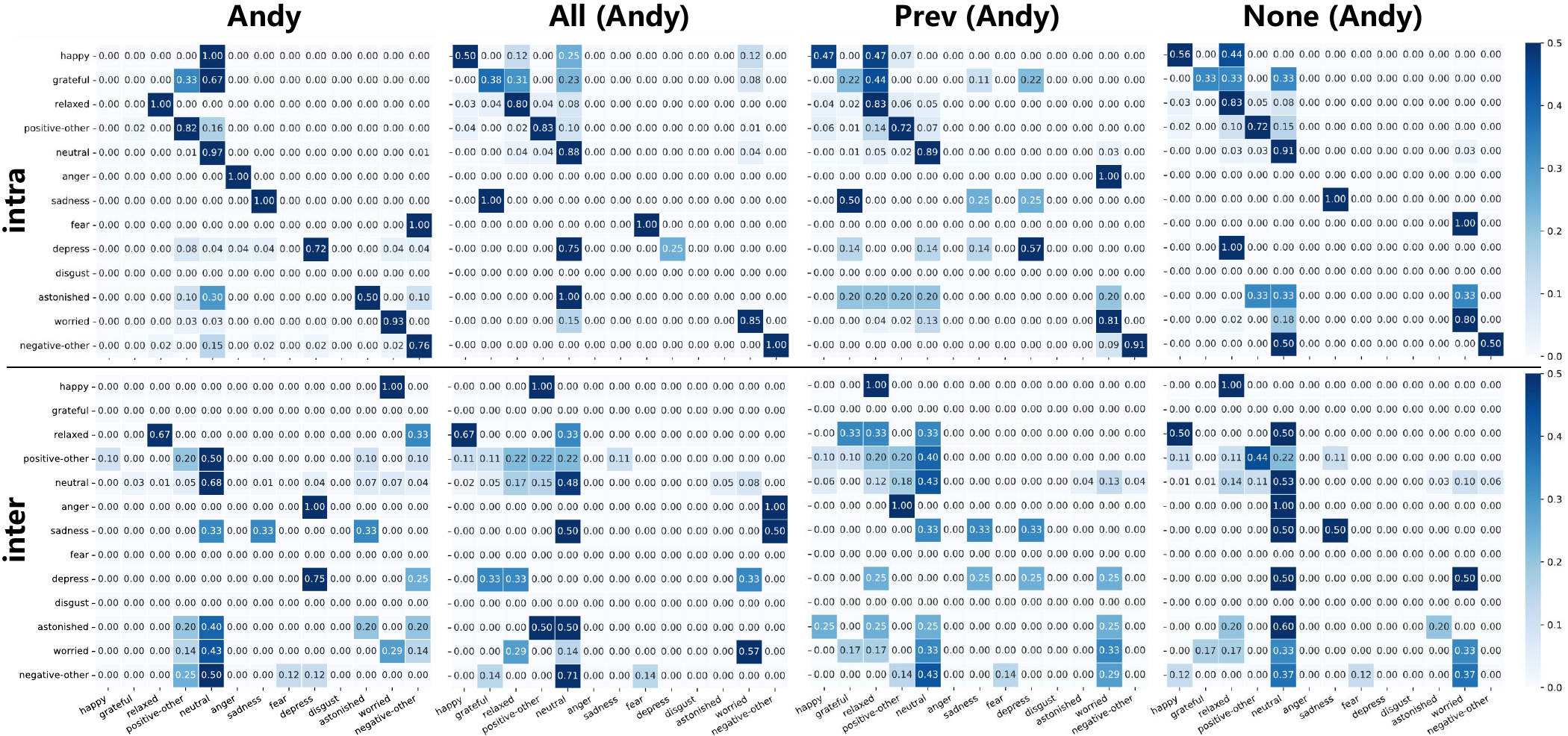} 
	\caption{Emotional transition matrices of Andy in \textit{Ode to Joy}. Groundtruth matrices are list in the first column. The rest columns are role-played by GPT-5-chat with different types.} 
	\label{fig:case_dynamic_emo}
\end{figure*}

\begin{figure}[t]
	\centering
	\includegraphics[width=\columnwidth]{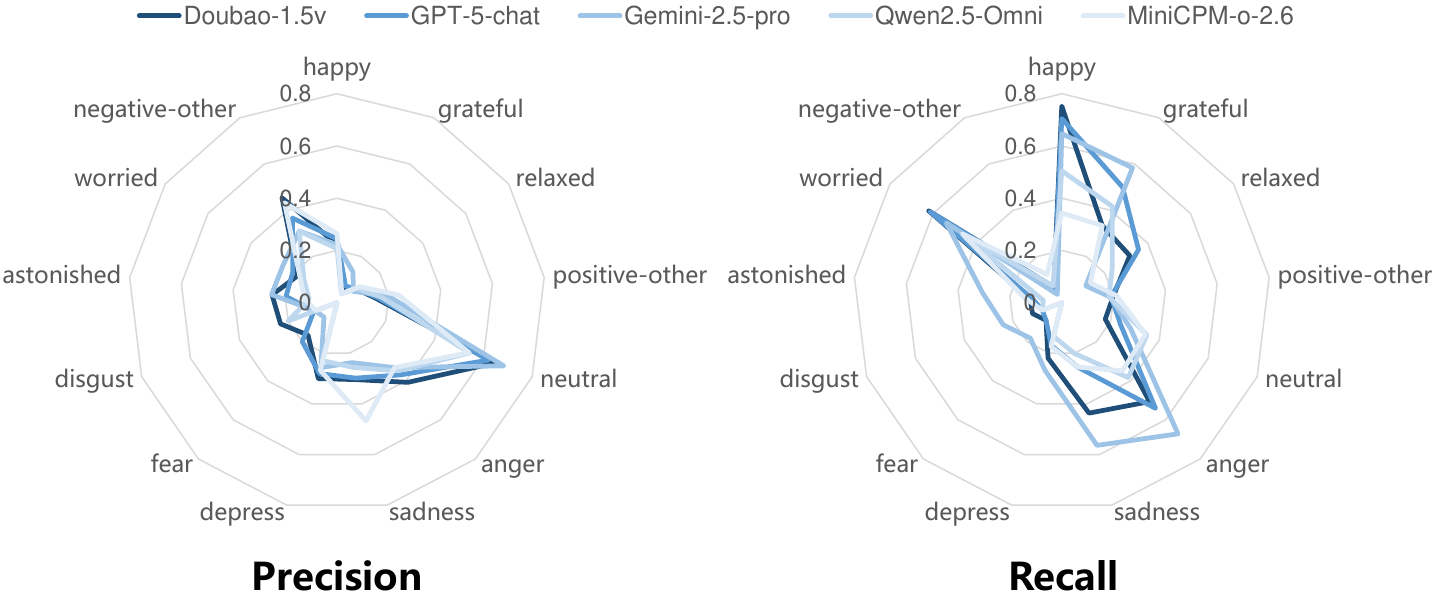} 
	\caption{The visualization of \textit{Recall} and \textit{Precision} scores of each emotions when calculating $MEC_{lower}$ on all models with $type=All$.}
	\label{fig:case_MEC_f1}
\end{figure}

Table~\ref{tab:train results} reports the evaluation results of {\ProposedFrameworkMERRY} on \textit{MiniCPM-o-2.6(8B)} trained on different role-playing datasets, including \textit{MMRole} \cite{dai2025mmrolecomprehensiveframeworkdeveloping}, \textit{OmniCharacter} \cite{zhang2025omnicharacterimmersiveroleplayingagents}, and {\ProposedDatasetMERRY} (Ours). We observe that:
\textbf{training on synthetic datasets would reduce emotional consistency.} 
Specifically, \textit{OmniCharacter} and \textit{MMRole} are both synthetic multimodal datasets, while {\ProposedDatasetMERRY} (Ours) are extracted from real multimodal data source. 
Model trained on \textit{MMRole} exhibits performance drop on $MEC_{upper}$, $CEC$, $EDD$, and $RCD$, %
while for \textit{OmniCharacter} the performance drops on $MEC$, $CEC$, $EDD_{inter}$, $RCD_{intra}$, and $ED$.
The decrease on \textit{OmniCharacter} is more severe than that on \textit{MMRole}, which can be due to the fixed emotion list in \textit{OmniCharacter}.
By contrast, \textbf{training on {\ProposedDatasetMERRY} (Ours) could improve emotional consistency}, reaching the best performance on $MEC$, $CEC$, $ED$.
However, although the performance improves on RC after training, model still potentially struggles with the large number of roles compared to \textit{MMRole} and \textit{OmniCharacter}, which contain 85 and 20 roles with a similar or larger sample size, respectively.
Therefore, role-related scores of our trained baseline, involving $EDD$, $RCD$, and $RC$, are less than satisfactory, which also aligns with the results of EmoCharacter \cite{feng-etal-2025-emocharacter}, indicating that \textbf{simple fine-tuning method suffers from poor role generalization}.
In summmary, synthetic data is easy to obtain and can be scaled up rapidly, which is helpful to a certain extent for quickly improving RC.
However, the alignment between synthetic and real emotions is relatively poor, which will lead to a decline in EC after training.

\subsection{Analysis}
\label{ablation_study}

To further explore the reasons and phenomena behind digital experimental results, we conduct in-depth analysis in this section.

\begin{figure*}[t]
	\centering
	\includegraphics[width=\textwidth]{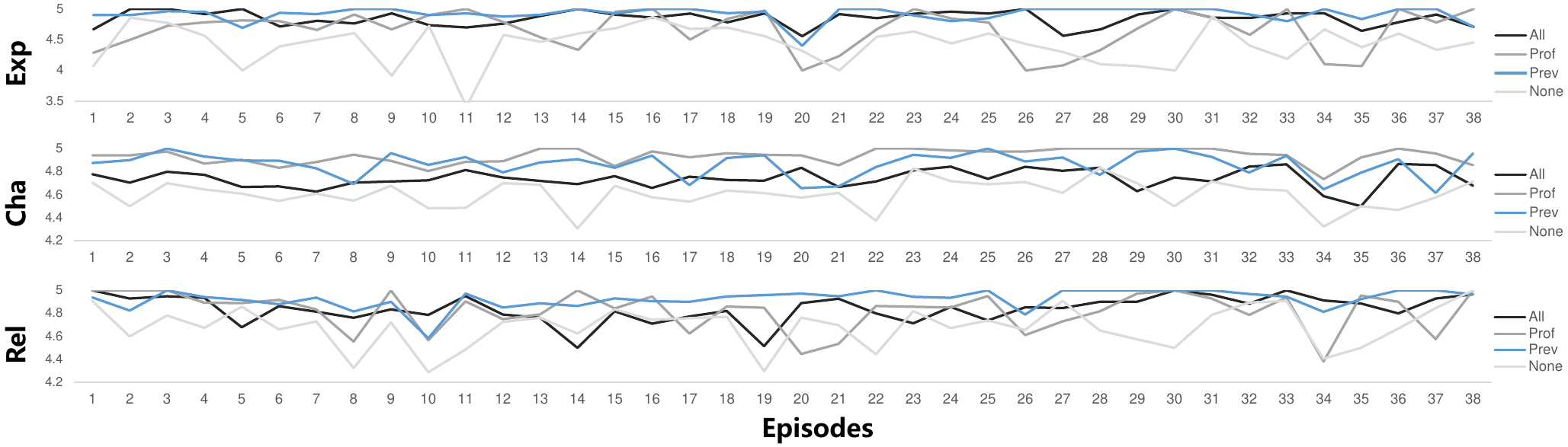} 
	\caption{Average of RC scores of Gemini-2.5-pro based on different episodes of \textit{Ode to Joy}. Notably, \textit{Ode to Joy} is a 2-series TV. Episodes 1 to 18 are from Series 1, while episodes 19 to 38 are from Series 2.} 
	\label{fig:case_RC}
\end{figure*}

\subsubsection{Why Emotional Inconsistency}
\textbf{Existing models suffer from emotional templatization and simplification, exhibiting positive-bias and performance bottleneck in fine-grained negative emotions.}
As shown in Figure~\ref{fig:case_MEC_f1}, models exhibit a tendency toward the overuse of specific emotions, resulting in high recall but low precision, including \textit{happy}, \textit{grateful}, \textit{relaxed}, \textit{anger}, \textit{sadness}, and \textit{worried}.
This suggests that existing models have following issues:
(1) \textbf{Positive Biases}: After RLHF, models tend to be overly ingratiating, habitually adopting a veneer of gratefulness or happiness regardless of contextual appropriateness;
(2) \textbf{Emotional Templatization and Simplification}: Common basic emotions such as happiness, anger and sadness are easily adopted by models as a one-size-fits-all template for negative expressions. 
This results in misrepresentation of subtle emotions as blunt, template-based ones.
For instance, rendering a state of lacking spirit and motivation (\textit{depress}) as an intense feeling of sorrow and grief (\textit{sadness}).

Conversely, models exhibit restraint in expressing \textit{neutral} emotion, characterized by high precision but low recall.
This indicates that in multimodal role-playing scenarios, models are easily induced to express emotions with specific tendency, finding it difficult to maintain neutrality.
Finally, categories such as \textit{depress}, \textit{fear}, \textit{disgust}, and \textit{astonished} suffer from low recall and low precision simultaneously.
These dimensions represent the current bottlenecks of existing models, reflecting a lack of both the sensitivity and the fidelity required to express complex or visceral human emotions.


\subsubsection{Why Emotional Transitions Inconsistency}
\textbf{Existing models struggle with $EDD$, which ostensibly measures the consistency of emotional transitions but actually reflecting EC, RC, and human-likeness simultaneously.}
As shown in the first column (GT) of Figure~\ref{fig:case_dynamic_emo}, Andy tends to shift toward neutral or negative emotions during expression, which is consistent with her personality characterized by rationality, pessimism and realism. 
This tendency is more pronounced in the intra-turn matrix, further supporting that \textbf{intra-turn dynamic emotional transitions are often more closely related to character personality traits}, whereas inter-turn transitions additionally depend on varying contextual information.
In RPAs, because \textit{Prev} and \textit{None} do not incorporate profile constraints, they are more inclined to shift toward positive emotions, which leads to lower $EDD$ scores.
Although \textit{All}, which includes profile information, is still influenced by the model's general bias toward positivity, this effect is substantially mitigated, and the emotional transitions gradually align more closely with the ground-truth patterns.
However, \textbf{contrary to the sparse real-world emotion transitions, models potentially tend to exhibit more scattered emotional shifts, resulting in a lower degree of human-likeness}.
This also indicates that $EDD$ measures not only the consistencies of roles and emotions, but also the degree of human-likeness, with considerable scope for enhancement across various models.


%



\subsubsection{The Effect of Role-related Information}
As shown in Figure~\ref{fig:case_RC}, lacking \textit{Prev} leads to performance decline on $Exp$ and $Rel$ across various episodes, while the lack of \textit{Prof} leads to lower $Cha$.
This further demonstrates that \textit{Prev} provides important previous information for grounding time-varying knowledge, including experience and relationship, while \textit{Prof} is particularly beneficial for exhibiting the intended character traits.

\subsection{Effectiveness of {\ProposedFrameworkMERRY}}
Since, LLMs introduce subjectivity,
we recruit three graduate students to conduct human agreement experiments on:
\begin{enumerate*}[label=(\arabic*), itemsep=0pt, topsep=0pt, parsep=0pt, partopsep=0pt] 
	\item \textbf{Dataset Quality}: Human raters are asked to score the semantic consistency between three synthesized descriptions and real-world videos over 100 randomly selected samples in the unchanged subset where the descriptions are identified as \enquote{correct} by LLMs, on a scale of 1 to 5. 
	\item \textbf{Effectiveness of LLM-based Metrics}: Human raters are asked to identify bidirectional evidences on three RC metrics with the same instructions over 100 randomly selected samples in evaluation results. Same mapping table as Table~\ref{tab:rc_map_rule} is applied and scores from three raters are averaged, thereby creating groundtruth scores for subsequent comparison.
	\item \textbf{Effectiveness of ERC experts}: Human raters are asked to recognize emotions for each utterance over 100 randomly selected samples in evaluation results, with the same instructions as ERC experts. Notably, two additional human raters are recruited to maintain the same number of experts as {\ProposedFrameworkMERRY}. Emotions with less than four votes of five human raters are classified as ambiguous emotions.
\end{enumerate*}

\begin{table}[t]
	\centering
	\caption{Results of human agreement experiment on description quality. $\alpha$ refers to Krippendorff's alpha. $\bar{s}$ refers to the average score.} 
	\label{tab:human_agreement_data}
	\resizebox{\linewidth}{!}{%
		\begin{tabular}{@{}lccc@{}}
			\toprule
			\multicolumn{1}{c}{} & facial expression & body movement & speech prompt \\ \midrule
			$\bar{s}$                                      & 4.348                & 4.323            & 4.719            \\
			$\alpha$                    & 0.585              & 0.539          & 0.571          \\ \bottomrule
		\end{tabular}%
	}
\end{table}

Since the scoring results are ordinal data (except for nominal ERC results) evaluated by multiple raters and there are invalid values in RC metrics, we use Krippendorff's alpha \cite{Krippendorff2011ComputingKA} as metric of agreement.
Average scores in Table~\ref{tab:human_agreement_data} demonstrate substantial semantic consistency (greater than 4) between synthesized descriptions and real videos, with moderate confidence (from 0.4 to 0.6).
Results in Table~\ref{tab:human_agreement_metric} also indicate that under the proposed approach, LLM-based metircs have reached to substantial agreement (from 0.6 to 0.8) with human raters on all proposed metrics.
During ablation, the results further demonstrate that depending exclusively on either a single model or subjective scoring approach with explanations can introduce a certain degree of inconsistency.
Finally, results in Table~\ref{tab:human_agreement_ERC} exhibit that human raters and ERC experts have achieved substantial agreement, validating the effectiveness of proposed evaluation setup.
Notably, since role profiles, previous information, and corrected descriptions have already been manually verified during construction, we omit redundant agreement experiments on these contents.

\begin{table}[]
	\centering
	\caption{Krippendorff's alpha between human and different evaluator with different evaluation type. Best results are in \textbf{bold}.}
	\label{tab:human_agreement_metric}
	\resizebox{\columnwidth}{!}{%
		\begin{tabular}{@{}ccccc@{}}
			\toprule
			\textbf{Evaluator} & \textbf{Type} & \textbf{Exp} & \textbf{Cha} & \textbf{Rel} \\ \midrule
			Avg. (Ours) & \multirow{3}{*}{\begin{tabular}[c]{@{}c@{}}finding\\ bidirectional\\ evidences\end{tabular}}    & \textbf{0.745} & \textbf{0.619} & \textbf{0.721} \\
			GPT-5.2            &               & 0.631        & 0.552        & 0.619        \\
			Doubao-1.6         &               & 0.503        & 0.443        & 0.492        \\ \midrule
			Avg.        & \multirow{3}{*}{\begin{tabular}[c]{@{}c@{}}subjective scoring\\ with explanations\end{tabular}} & 0.558          & 0.386          & 0.482          \\
			GPT-5.2            &               & 0.490        & 0.354        & 0.407        \\
			Doubao-1.6         &               & 0.367        & 0.238        & 0.339        \\ \bottomrule
		\end{tabular}%
	}
\end{table}


\begin{table}[]
	\centering
	\caption{Krippendorff's alpha between human and ERC experts across different modalities.} 
	\label{tab:human_agreement_ERC}
	\resizebox{\columnwidth}{!}{%
		\begin{tabular}{@{}lcccc@{}}
			\toprule
			\multicolumn{1}{c}{} & facial expression & body movement & speech prompt & fusion \\ \midrule
			$\alpha$  & 0.696 & 0.656 & 0.709 & 0.721 \\ \bottomrule
		\end{tabular}%
	}
\end{table}

\section{Discussion and Conclusion}
Current researches entangle semantic assessment with modality generation, thereby leading to ambiguity in issue attribution and high dependence on human judgement.
To address the gap,
we propose {\ProposedFrameworkMERRY}, a semantically decoupled evaluation framework for assessing \textbf{M}ultimodal \textbf{E}motional and \textbf{R}ole consistencies of \textbf{R}ole-pla\textbf{y}ing agents, and {\ProposedDatasetMERRY}, a multi-turn multimodal role-playing emotional interaction dialogue dataset, and conduct extensive experiments. 
Highlighted findings are as follows:
{\Findings}


Due to the limitations of qualified human resource for quality control and human assessment, this paper primarily focus on Chinese. 
To support more research, we are planning to open-source all the codes in this article, including the data construction pipeline and the proposed evaluation framework {\ProposedFrameworkMERRY}, thereby enabling the reproduction in other languages. 
\bibliographystyle{IEEEtran}
\bibliography{anthology,custom}

\end{document}